\def\paperID{**}
\def\confName{CVPR}
\def\confYear{2025}

\def\paperTitle{
Pre-training with 3D Synthetic Data:\\Learning 3D Point Cloud Instance Segmentation from 3D Synthetic Scenes
}

\def\authorBlock{
    Daichi Otsuka$^{\dag}$ \ \ Shinichi Mae$^{\dag}$ \ \ Ryosuke Yamada$^{\ddag}$ \ \ Hirokatsu Kataoka$^{\ddag,\diamondsuit}$ \\
    \dag TICO-AIST Cooperative Research Laboratory for Advanced Logistics (ALlab) \\
    \ddag National Institute of Advanced Industrial Science and Technology (AIST) \\ $\diamondsuit$ Visual Geometry Group, University of Oxford \\
    {\tt\small \ ootsuka.da@aist.go.jp}
}

\newif\ifreview 
\newif\ifarxiv 
\newif\ifcamera \newcommand{\cameraready}{\cameratrue}
\newif\ifrebuttal

\cameraready

\pdfoutput=1
\documentclass[10pt,twocolumn,letterpaper]{article}

\ifreview \usepackage[review]{cvpr} \fi
\ifarxiv \usepackage[pagenumbers]{cvpr} \fi
\ifrebuttal \usepackage[rebuttal]{cvpr} \fi
\ifcamera \usepackage{cvpr} \fi

\usepackage{xr-hyper}

\makeatletter
\newcommand*{\addFileDependency}[1]{
  \typeout{(#1)}
  \@addtofilelist{#1}
  \IfFileExists{#1}{}{\typeout{No file #1.}}
}

\makeatother

\definecolor{cvprblue}{rgb}{0.21,0.49,0.74}
\usepackage[pagebackref,breaklinks,colorlinks,allcolors=cvprblue]{hyperref}
\usepackage[capitalize]{cleveref}
\crefname{section}{Sec.}{Secs.}
\crefname{table}{Table}{Tables}
\crefname{figure}{Fig.}{Figs.}

\ifarxiv \crefname{appendix}{App.}{Apps.}
\else \crefname{appendix}{Suppl.}{Suppls.} \fi

\usepackage{graphicx}   
\usepackage{amsmath}    
\usepackage{amssymb}    
\usepackage{booktabs}
\usepackage{times}
\usepackage{microtype}
\usepackage{epsfig}
\usepackage{caption}
\usepackage{float}
\usepackage{placeins}
\usepackage{color, colortbl}
\usepackage{stfloats}
\usepackage{enumitem}
\usepackage{tabularx}
\usepackage{xstring}
\usepackage{multirow}
\usepackage{xspace}
\usepackage{url}
\usepackage{subcaption}
\usepackage{xcolor}
\usepackage[hang,flushmargin]{footmisc}
\usepackage{comment}

\ifcamera \usepackage[accsupp]{axessibility} \fi

% ...

\ifarxiv  \fi

\newcommand{\R}[1]{{%
    \textbf{%
        \ifstrequal{#1}{1}{\textcolor{red}{R#1}}{%
        \ifstrequal{#1}{2}{\textcolor{blue}{R#1}}{%
        \ifstrequal{#1}{3}{\textcolor{magenta}{R#1}}{%
        \ifstrequal{#1}{4}{\textcolor{teal}{R#1}}{%
                           \textcolor{cyan}{R#1}%
        }}}}%
    }%
}}

\begin{document}
\title{\paperTitle}
\author{\authorBlock}
\maketitle

\begin{abstract}
In the recent years, the research community has witnessed growing use of 3D point cloud data for the high applicability in various real-world applications. By means of 3D point cloud, this modality enables to consider the actual size and spatial understanding. The applied fields include mechanical control of robots, vehicles, or other real-world systems. Along this line, we would like to improve 3D point cloud instance segmentation which has emerged as a particularly promising approach for these applications. However, the creation of 3D point cloud datasets entails enormous costs compared to 2D image datasets. To train a model of 3D point cloud instance segmentation, it is necessary not only to assign categories but also to provide detailed annotations for each point in the large-scale 3D space. Meanwhile, the increase of recent proposals for generative models in 3D domain has spurred proposals for using a generative model to create 3D point cloud data. In this work, we propose a pre-training with 3D synthetic data to train a 3D point cloud instance segmentation model based on generative model for 3D scenes represented by point cloud data. We directly generate 3D point cloud data with Point-E for inserting a generated data into a 3D scene. More recently in 2025, although there are other accurate 3D generation models, even using the Point-E as an early 3D generative model can effectively support the pre-training with 3D synthetic data. In the experimental section, we compare our pre-training method with baseline methods indicated improved performance, demonstrating the efficacy of 3D generative models for 3D point cloud instance segmentation.
\end{abstract}

\begin{figure*}[t]
    \centering
    \includegraphics[width=1.0\linewidth]{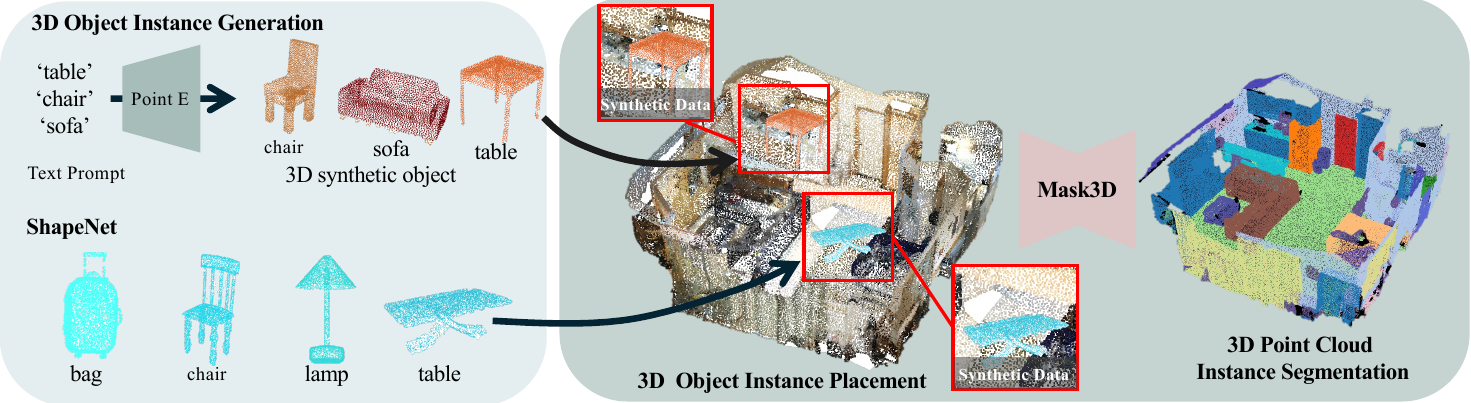}
    \vspace{-10pt}
    \caption{\textbf{3D scene expansion utilising a single 3D generative model.} In the present paper, we employ Point-E as a 3D object generative model, to automatically synthesize 3D object instances. By placing generated 3D object instances in existing 3D scene data (e.g., ScanNetV2), the 3D scene data is effectively extended with generated 3D object instances. In the 3D generative pre-training pipeline, we use Point-E to generate 3D object instances and expand ScanNetV2 by randomly placing them into 3D scenes. The center of gravity coordinates of the 3D objects and scenes are aligned during insertion. Then random noise is added to the coordinates of the center of gravity of the 3D objects. In the pre-training phase, we assign Mask3D as a 3D point cloud instance segmentation model and pre-train the 3D point dloud instance segmentation model on the expanded 3D scene dataset. We additionally fine-tune the pre-trained Mask3D model on S3DIS dataset. We successfully show the effectiveness of `3D generative pre-training', a pre-training method for 3D point cloud instance segmentation from a single generative model.}
    \label{fig:dataset_pipline}
    \vspace{-15pt}
\end{figure*}

\section{Introduction}

Improvements in the performance of sensor technology (e.g., LiDAR) and virtual reconstruction methodologies for measuring the real world in 3D have spurred growing use of 3D point cloud data for mechanical control of robots, vehicles, and other systems. Among various existing techniques, the method of 3D point cloud instance segmentation has been used to be applied input for tasks such as automated driving and robot actuation, due to its ability to infer pixel-specific position information for individual object instances in 3D point clouds~\cite{armeni20163d, akizuki2018tactile}.

Compared to conventional instance segmentation in 2D images, 3D point cloud instance segmentation from an input with point clouds uniquely considers spatial attributes such as depth, orientation, and scale, making it particularly effective in tasks requiring precise 3D spatial understanding. For example, such as an application for autonomous driving and autonomous mobile robots, accurate instance segmentation of individual objects, pedestrians, and other obstacles in 3D space is crucial to determine collision-free paths, thereby directly influencing safety and efficiency. Similarly, in robotic manipulation tasks, precise 3D point cloud instance segmentation at each object allows robots to accurately grasp, move, and manipulate various objects within complex environments.

However, 3D point cloud instance segmentation has conventionally required object labels, and meanings for specific 3D points, assigned to 3D point clouds by human operators capable of accounting for rotations and other transformations in 3D point cloud data. Compared to 2D images, this entails enormously greater personnel costs for dataset construction and explains why large-scale 3D point cloud datasets with annotations at each point are relatively rare. In practice, annotators must not only semantic label each object instance but also precisely divide boundaries across thousands or even millions of spatial points. Such human annotation complexity drastically limits the scalability in the dataset construction of 3D point clouds. This issue imposes a bottleneck for advancing 3D point cloud instance segmentation in both academic and industry purposes.

Recently, on the other hand, we have witnessed significant advances in generative models, especially image generation with diffusion models~\cite{rombach2022high, saharia2022photorealistic, ramesh2022hierarchical}. The generative models have been also employed in 3D generative models such as Point-E~\cite{nichol2022point}. At the same time, the key techniques will induce a novel perspective for addressing the bottleneck due to the dataset construction in the context of 3D recognition domain. These generative models enables to synthesize labeled 3D point cloud data from minimal input (e.g., texts, words), potentially bypassing manual annotation efforts in pre-training phase. By leveraging the ability of generative models to produce synthetic yet pseudo 3D point cloud data, it is now feasible to generate large-scale, diverse training 3D point cloud datasets with detailed instance/point-level semantic annotations automatically provided during the data generation process from a single 3D generative model.

The pre-training with 3D synthetic data with a 3D generative model, capable of robust 3D point cloud instance segmentation in 3D point clouds would be of great significance for a well-deserved solution to 3D dataset preparation. Building upon the potential, our work of pre-training with 3D synthetic data specifically explores the utilization of a single 3D generative model in the form of Point-E for directly constructing labeled data suitable for a model training of 3D point cloud instance segmentation. Through extensive experimental results, we demonstrate how pre-training with 3D synthetic data from a text input can match or even surpass segmentation performance obtained through traditionally and manually annotated datasets. Our approach significantly reduces the practical barriers to developing robust and scalable 3D point cloud instance segmentation solutions, paving the way toward broader adoption and accelerated research progress in real-world applications.

\section{Related work}

As vision foundation models, general-purpose instance-segmentation methods for 2D images include semi-automatic data-collection approaches such as segment anything~\cite{kirillov2023segment}, which applies an initial segmentation scheme to an arbitrary pre-trained model, then prepares training data with only simple sanity checks contributed by humans. The resulting trained vision foundation model, the segment anything model (SAM), is said that capable of general-purpose segmentation of real-world object instances.
In the recent years, we have also witnessed numerous proposals of generative models for text-to-image models, including the series of Stable Diffusion (e.g., \cite{rombach2022high}) and DALL·E (e.g., \cite{ramesh2022hierarchical}). Generative models for automatically creating image datasets and building trained models have been shown the effectiveness in 2D image domains~\cite{fan2024scaling, sariyildiz2023fake}. Therefore, we have great potentials to assign such generative models for training 3D recognition. Unlike the 2D image generation, the approach considered optimal for 3D point cloud data is to exclude the complex color combinations and textures found in the foregrounds and backgrounds of 2D images, focusing instead on 3D point cloud shapes and orientations. This ensures that generative models for 3D point clouds can readily reproduce data obtained by scanning real-world objects while simultaneously mitigating the impact of defects and distortions commonly encountered in real-world scanned data. By adding intentional defects to generated 3D point cloud data, or via similar strategies, these techniques can also be used to create datasets incorporating diversity. Generative models proposed for 3D point clouds include Point-E~\cite{nichol2022point}, which reconstructs 3D shapes under the representation of point cloud data from input text, and generated 3D data from a single mathematical formula~\cite{yamada2022point,PC-Perlin,VG-Fractal}, an approach based on fractal geometry or Perlin noise which despite including no real-world objects in its pre-training dataset nonetheless achieved a high performance for 3D object recognition at the time of its proposal, demonstrating the effectiveness of auto-generated 3D point cloud data. Thus, the application of generative models for 3D point cloud data is a promising approach for creating high-quality 3D point cloud instance segmentation models.

In this work, we use Point-E as a 3D point cloud generative model to create a general-purpose instance segmentation model for 3D point clouds. The use of generative model for this purpose offers several advantages, including (1) auto-generation of 3D point cloud data with immediate auto-creation of pre-trained models, (2) the ability to create training datasets customized for specific tasks, and (3) a solution to the problem of insufficient 3D point cloud data thanks to our ability to auto-generate theoretically infinite datasets with associated training data.

\section{Pre-training with 3D synthetic data}
We propose a concept of a pre-training with 3D synthetic data. In this paper, we conduct an 3D object instance generation and the generated objects can be employed for 3D dataset expansion. Here, we try to automatically generate 3D object instances for training a 3D point cloud instance segmentation model. The 3D dataset generation framework we propose is outlined schematically in Figure~\ref{fig:dataset_pipline}.

\noindent \textbf{Base 3D scene.} The generated 3D object instances can be assigned to all scenes in the ScanNetV2~\cite{dai2017scannet} dataset. ScanNetV2 is one of the most used 3D datasets that collects and reconstructs 3D scenes and object instances in various indoor spaces. By simultaneously acquiring RGB images and depth information, each scene is captured from multiple viewpoints and represented as integrated 3D point cloud data. Moreover, each 3D object instance within the 3D scene is manually annotated with instance-level semantic labels, detailing the shape and position of each object instance. Nowadays, it has been widely used as foundational data for analyzing the 3D structure and object arrangement in 3D indoor environments.

\noindent \textbf{3D object instance generation.} We begin by data generation for 3D point cloud object instances using the Point-E. We then extend this 3D point cloud dataset by using ScanNetV2 as described in base 3D scene, a pre-existing 3D point cloud dataset, to position auto-generated 3D point cloud object instances within various scenes reflecting indoor environments.
In each 3D scene on ScanNetV2, we extend the 3D scenes by randomly selecting generated 3D object instances from a single generative model (and additional 3D models). The text prompts consist of a single word (e.g., table, chair, sofa). Furthermore, to maintain simplicity in 3D scene expansion, only up-to two object instances are projected in each scene. In order to achieve more realistic 3D scene synthesis, we also synthesize mixed data by randomly selecting from ShapeNet~\cite{chang2015shapenet}, a collection of 3D CAD data.

\noindent \textbf{3D object instance placement.} We now describe our method for positioning auto-generated 3D object instances within an indoor-environment 3D scene. First, we compute the center of gravity (COG) for the object instance within the 3D scene represented by 3D point clouds for the 3D indoor environment. Next, we overlay the COG of object instance atop the COG coordinates of the 3D point cloud data for the 3D indoor environment. Finally, we add random noise to the COG coordinates of the 3D object instance. The result of this procedure is to position the auto-generated 3D object instance within the indoor-environment scene.

\noindent \textbf{Pre-training with 3D synthetic data and fine-tuning from the pre-trained model.} During the pre-training stage on the expanded ScanNetV2 with Point-E, we assign Mask3D~\cite{schult2023mask3d} as a 3D point cloud instance segmentation model and train the model on the expanded 3D scenes with ScanNetV2. The pre-trained Mask3D model is further fine-tuned on the S3DIS dataset~\cite{s3dis}. Our results successfully demonstrate the effectiveness of ``pre-training with 3D synthetic data,'' a method that leverages a single generative model for 3D point cloud instance segmentation.

\section{Experiments}
In the experimental section, we perform pre-training extended by a single 3D generative model (utilizing Point-E). We also verify the effectiveness of our proposed method by comparing it with a training from scratch and a simple pre-training with original ScanNetV2.

\subsection{Pre-training and fine-tuning}
We employ ScanNetV2 as the base pre-training dataset and S3DIS as the fine-tuning dataset in the experiments. For the comparison, we evaluate training from scratch (without pre-training), pre-training on plain ScanNetV2, and pre-training on ScanNetV2 expanded using Point-E.

Here, ScanNetV2 is a 3D scene dataset scanned from 1,513 indoor scenes and contains 50 categories. In our experiments, the number of instances in the plain ScanNetV2 pre-training is 48,698, while our extended ScanNetV2 has 51,100 instances. The increase in generated 3D object instances is relatively small (+2,402 ojbect instances) because we aimed to validate whether minimal object instance additions can improve the pre-training effect in a 3D point cloud instance segmentation.

On the other hand, the fine-tuning phase on S3DIS dataset is a 3D scene dataset scanned from the interiors of six different buildings at Stanford University. Typically, a model fine-tuning is performed on five areas (excluding Area 5) and validating is conducted on Area 5. The same procedure is followed in our experiments. S3DIS consists of 271 scenes and 13 categories.

\subsection{Training details and setup}

The model used in the experiment is Mask3D, a widely assigned and sophisticated model for 3D point cloud instance segmentation. In the pre-training phase, the model was trained with a batch size of 4, a learning rate of 0.0001, a voxel size of 30 mm, and for 600 epochs.

After the pre-training phase, we additionally fine-tune Mask3D on the S3DIS dataset for 3D point cloud instance segmentation. The hyperparameters during fine-tuning remained the same as during pre-training following the conventional approaches. In our experiments, the pre-training and fine-tuning are conducted using a single NVIDIA A6000 GPU.

\subsection{Experimental results}
In this subsection, we primarily compare the results of 3D point cloud instance segmentation among three different pre-training methods including training from scratch, plain ScanNetV2 pre-training, and proposed expanded ScanNetV2 pre-training. Table~\ref{tab:main_result} shows the results of fine-tuning results on S3DIS with each of the pre-trained Mask3D models.

At the beginning, compared to training from scratch, the results with plain ScanNetV2 pre-training show a significant improvement, an increase of 14.6 points in average precision (AP; 48.4 vs. 33.8). Furthermore, when ScanNetV2 is pre-trained on the dataset expanded with 3D object instances generated primarily by Point-E, an additional improvement of 2.9 points in AP (51.3 vs. 48.4) is further observed. This indicates that even a minimal expansion, with a maximum of two additional 3D object instances per 3D scene on ScanNetV2, can lead to notable accuracy gains.

Additionally, Table~\ref{tab:class_result} presents the AP for each object instance category. Despite the minimal 3D object instance expansion with up-to two object instances per 3D scene, our method records the highest AP among the three approaches in 7 out of 13 object instance categories (chair, sofa, beam, column, bookcase, board, and clutter). In particular, the 3D generative pre-training is effective for relatively small objects, as evidenced by the improvements for beam (from 4.3 to 25.9) and board (from 69.4 to 76.4).

\begin{table}[t]
    \centering
    \caption{~\textbf{Results on 3D point cloud instance segmentation.} We show the results of training from scratch, pre-training with ScanNetV2 and expanded ScanNetV2 (ours). These pre-trained models are additionally fine-tuned on S3DIS to validate in average precision (AP), and AP$_{50}$ with IoU 0.50 and AP$_{25}$ with IoU 0.25.}
    \vspace{-10pt}
    \begin{tabular}{l|c|c|c}
    \toprule
    \multirow{2}{*}{Pre-training method} & \multicolumn{3}{c}{S3DIS (Area 5)} \\
     & AP & AP$_{50}$ & AP$_{25}$ \\
    \midrule
    Training from scratch & 33.8 & 45.6 & 54.6 \\
    ScanNetV2 & 48.4 & 61.5 & 69.5 \\
    \rowcolor[gray]{0.9} ScanNet w/ 3D synthetic data & \textbf{51.3} & \textbf{63.3} & \textbf{71.8} \\
    \bottomrule
    \end{tabular}
    \label{tab:main_result}
\end{table}

\begin{table}[t]
    \centering
    \caption{\textbf{AP at each S3DIS category.} We list the 12 different categories on S3DIS dataset. The gray meshes describe for the object instance categories for which the training results from our proposed pre-training method has the highest AP.}
    \vspace{-10pt}
    \begin{tabular}{r|c|c|c}
    \toprule
    \multirow{2}{*}{Object category} & \multicolumn{3}{c}{Pre-training method} \\
     & Scratch & ScanNet & Ours \\
    \midrule
    \rowcolor[gray]{0.9} Chair & 63.4 & 83.7 & \textbf{84.7} \\
    \rowcolor[gray]{0.9} Sofa & 3.8 & 61.7 & \textbf{65.2} \\
    Table & 10.9 & \textbf{30.7} & 30.6 \\
    Ceiling & \textbf{80.4} & 76.1 & 76.1 \\
    Floor & 94.4 & \textbf{98.5} & 97.0 \\
    Wall & 51.3 & \textbf{60.3} & 56.0 \\
    \rowcolor[gray]{0.9} Beam & 0.0 & 4.3 & \textbf{25.9} \\
    \rowcolor[gray]{0.9} Column & 10.8 & 14.8 & \textbf{17.1} \\
    Window & 32.5 & 41.7 & \textbf{45.0} \\
    Door & 6.3 & \textbf{40.7} & 36.6 \\
    \rowcolor[gray]{0.9} Bookcase & 15.7 & 20.8 & \textbf{26.9} \\
    \rowcolor[gray]{0.9} Board & 47.6 & 69.4 & \textbf{76.4} \\
    \rowcolor[gray]{0.9} Clutter & 22.4 & 27.2 & \textbf{29.0} \\
    \bottomrule
    \end{tabular}
    \label{tab:class_result}
\end{table}

\section{Conclusion and discussion}
We investigated the effectiveness of pre-training with 3D synthetic data in augmented training with a single 3D generative model. We utilised Point-E, a text-to-3D generative model, to perform object instance expansion of the 3D scene dataset to verify the pre-training effectiveness in 3D point cloud instance segmentation. To this end, we focus on the number of instances of the dataset used during pre-training. 

Here, the total number of instances used in plain ScanNetV2 was 48,698, while our expanded ScanNetV2 contained 51,100 instances. Despite the increase being only 2,402, limited to a maximum of two object instances per scene, a significant improvement in AP was observed.

On the other hand, Point‑E is a text-to-3D generative model proposed in 2022 at an early stage in this topic. Although there were some of the  corrupted object generation, the proposed pre-trained Mask3D model demonstrates that Point‑E is sufficiently useful for 3D scene expansion in pre-training with 3D synthetic data.

In the paper, we employed Point‑E as a 3D generative model to directly generate 3D point clouds. By applying more recent and advanced text-to-3D models and converting their outputs into the 3D point cloud modality, it has a clear potential for even more advanced pre-training.

We also expect further improvements in performance rates by simply replacing the model with a more recent, advanced one and by adjusting the optimal number of object instances per 3D scene. In the near future, pre-training with 3D synthetic data may become sophisticated enough to complete 3D modality training without any real data, and for applications in robotics, autonomous driving, and beyond.

{\small
\bibliographystyle{ieeenat_fullname}
\bibliography{main}
}

\ifarxiv \clearpage \appendix \input{12_appendix} \fi

\end{document}